\documentclass[sigconf]{acmart}
\AtBeginDocument{%
  }
\usepackage{booktabs,tabularx,siunitx,makecell}
\usepackage{algorithm}
\usepackage{algpseudocode}
\usepackage{hyperref}
\hypersetup{
pdfcreator={My Tool},
pdfproducer={My Tool}
}

\copyrightyear{2025}
\acmYear{2025}
\setcopyright{cc}
\setcctype{by}
\acmConference[SIGSPATIAL '25]{The 33rd ACM International Conference on
Advances in Geographic Information Systems}{November 3--6,
2025}{Minneapolis, MN, USA}
\acmBooktitle{The 33rd ACM International Conference on Advances in
Geographic Information Systems (SIGSPATIAL '25), November 3--6, 2025,
Minneapolis, MN, USA}
\acmDOI{10.1145/3748636.3764183}
\acmISBN{979-8-4007-2086-4/2025/11}

\begin{document}

\title{MVeLMA: Multimodal Vegetation Loss Modeling Architecture for Predicting Post-fire Vegetation Loss}

\author{Meenu Ravi}
\email{ravim@vt.edu}
\affiliation{%
  \institution{Virginia Tech}
  \city{Alexandria}
  \state{Virginia}
  \country{USA}
}

\author{Shailik Sarkar}
\email{shailik@vt.edu}
\affiliation{%
  \institution{Virginia Tech}
  \city{Alexandria}
  \state{Virginia}
  \country{USA}
}

\author{Yanshen Sun}
\email{yansh93@vt.edu}
\affiliation{%
  \institution{Virginia Tech}
  \city{Alexandria}
  \state{Virginia}
  \country{USA}
}

\author{Vaishnavi Singh}
\email{vs699@georgetown.edu}
\affiliation{%
  \institution{Georgetown University}
  \city{Washington D.C.}
  \country{USA}
}

\author{Chang-Tien Lu}
\email{clu@vt.edu}
\affiliation{%
  \institution{Virginia Tech}
  \city{Alexandria}
  \state{Virginia}
  \country{USA}
}


\begin{abstract}
Understanding post-wildfire vegetation loss is critical for developing effective ecological recovery strategies and is often challenging due to the extended time and effort required to capture the evolving ecosystem features. Recent works in this area have not fully explored all the contributing factors, their modalities, and interactions with each other. Furthermore, most research in this domain is limited by a lack of interpretability in predictive modeling, making it less useful in real-world settings. In this work, we propose a novel end-to-end ML pipeline called MVeLMA (\textbf{M}ultimodal \textbf{Ve}getation \textbf{L}oss \textbf{M}odeling \textbf{A}rchitecture) to predict county-wise vegetation loss from fire events. MVeLMA uses a multimodal feature integration pipeline and a stacked ensemble-based architecture to capture different modalities while also incorporating uncertainty estimation through probabilistic modeling. Through comprehensive experiments, we show that our model outperforms several state-of-the-art (SOTA) and baseline models in predicting post-wildfire vegetation loss. Furthermore, we generate vegetation loss confidence maps to identify high-risk counties, thereby helping targeted recovery efforts. The findings of this work have the potential to inform future disaster relief planning, ecological policy development, and wildlife recovery management.
\end{abstract}




\maketitle
\begin{figure*}[t]
  \centering
  \includegraphics[width=1\textwidth]{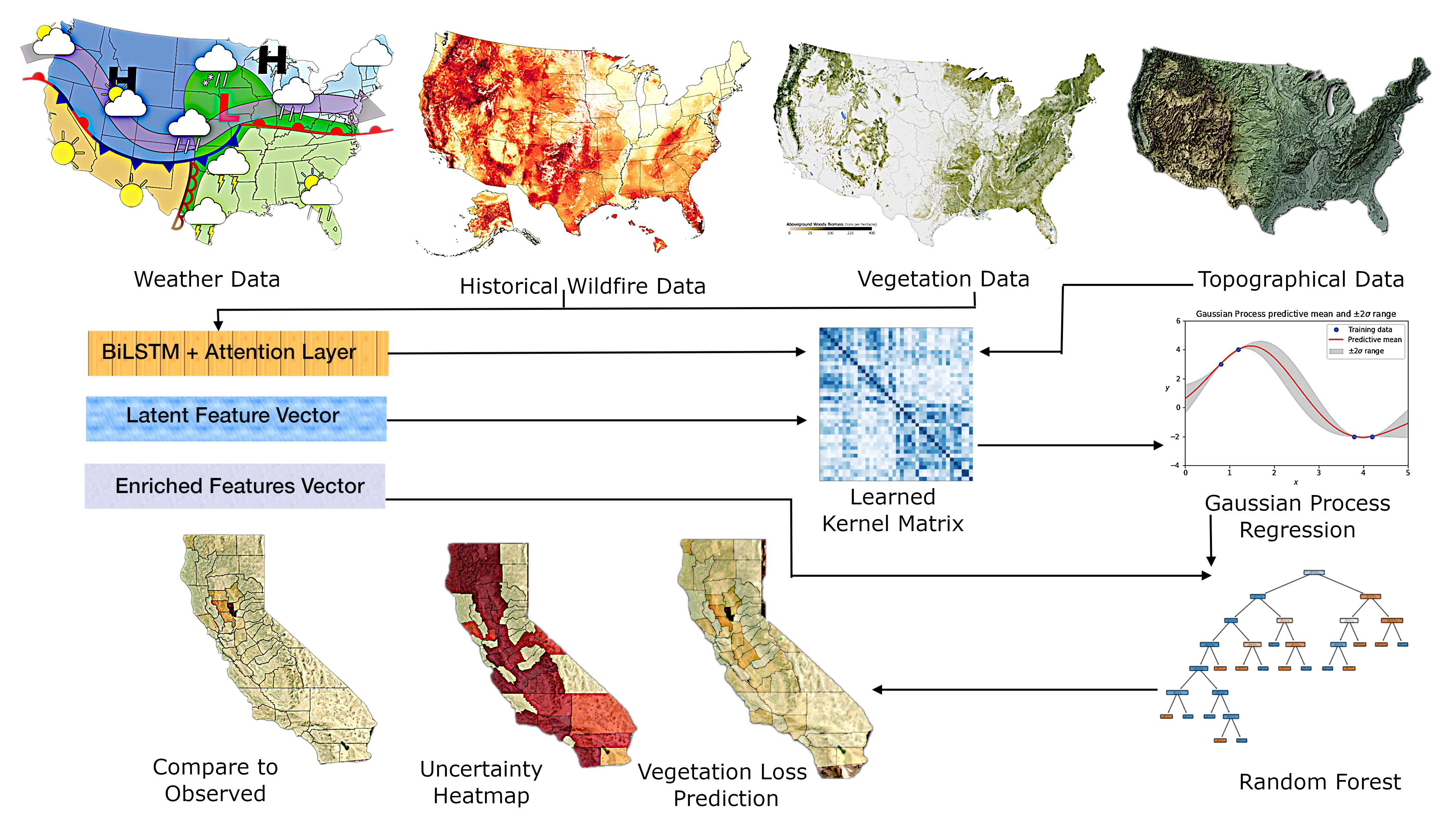} %
  \caption{ Overview of the proposed pipeline architecture for vegetation loss prediction, combining multimodal data sources, temporal encoding, uncertainty estimation, and predictive modeling.}
  \label{fig:model_pipeline}
\end{figure*}
\section{Introduction}
The prevalence of wildfires has increased significantly in recent years. Driven by climate change and other environmental factors, these natural disasters can disrupt the ecological balance, destroy wildlife habitats, devastate agricultural areas, and contribute to global warming by releasing harmful gases into the atmosphere \cite{deng_wildfire_2023, kabir_wildfire_2024, gerard_wildfirespreadts_2023, bolan_wildfires_2025,castel-clavera_disentangling_2022}. Identifying affected areas and monitoring land coverage are essential for ecological management and informing environmental and land-use policy decisions at various scales. To establish effective ecological recovery strategies and manage land area more effectively, it is critical to understand how fire affects vegetation and to identify spatiotemporal patterns in post-fire vegetation loss.

In recent years, there has been an increasing interest in predicting wildfire-related damage, particularly burned areas, wildlife loss, and infrastructure destruction \cite{priya_vegetation_2024}. The use of remote sensing data for vegetation health monitoring, specifically through indicators such as vegetation indices have become prevalent. Particularly, the Normalized Difference Vegetation Index (NDVI) is frequently used to evaluate vegetation health, study agricultural productivity, and monitor vegetation responses to climate-related events  \cite{priya_vegetation_2024, ferchichi_forecasting_2022, kartal_next-level_2024, wang_prediction_2025, vasilakos_lstm-based_2022, zhang_analyzing_2021, reddy_prediction_2018, ji_methods_2024, ebadati_rapid_2024, liu_forecasting_2023,dokhan_neural_2024, west_remote_2019}. 

Existing literature primarily focuses on time series forecasting approaches and machine learning (ML) models applied to remote sensing indices such as NDVI. Vegetation forecasting research has predominantly focused on leveraging meteorological variables (e.g., precipitation, temperature, and climate data), along with soil properties, crop characteristics, and remote sensing indices such as NDVI \cite{priya_vegetation_2024}. A wide range of frameworks have been developed to address spatial and temporal complexities involved in vegetation modeling. Particularly, deep learning algorithms such as Deep Neural Networks (DNN), Convolutional Neural Networks (CNN), Recurrent Neural Networks (RNN), and CNNLSTM have been used due to their ability to extract dependencies from large datasets. These methods have been applied to various use cases, including drought prediction, landslide detection, crop yield estimation, drought damage assessment, plant disease detection, and wildfire spread modeling \cite{bhowmik_multi-modal_2023, deng_daily_2025, jadouli_deep_2025, ferchichi_forecasting_2022, lin_forest_2023, kartal_next-level_2024, masrur_capturing_2024, reddy_prediction_2018, yao_spatio-temporal_2022}.
However, a clear research gap remains in forecasting vegetation loss specifically caused by wildfires. Additionally, limited work exists that leverages land cover features and variables beyond traditional meteorological inputs, such as solar radiation, fire-specific temporal data, and vapor pressure.

Traditional time-series forecasting models, while efficient for stationary time series, struggle to handle non-linear relationships present in multivariable and remote-sensing data. Similarly, traditional ML approaches can find it challenging to capture the spatiotemporal dynamics and high-dimensionality of such data. Furthermore, these works largely ignore interpretability through uncertainty quantification, which remains significant in real-world applications of such predictive tools.

To address these limitations, we propose an end-to-end, efficient, uncertainty-quantifying and scalable multimodal framework for predicting county-level vegetation loss following wildfire events, called MVeLMA.

Our approach focuses on modeling nine dynamic meteorological features along with twenty-seven enriched and static topographical features collected over multiple years. This framework combines deep neural recurrent networks for capturing temporal patterns, Gaussian Process Regression (GPR) for modeling spatial correlations that produce uncertainty-aware predictions, and an ensembled decision tree-based residual network to refine these predictions by incorporating static topographic and enriched features. To the best of our knowledge, no previous studies have implemented such a multimodal approach for wildfire-related vegetation loss prediction. Therefore, this study serves as a novel solution to the primary objective of our study, which is to predict vegetation loss following wildfire events. 
The key contributions of our work can be summarized as follows:
\begin{itemize}

\item \textbf{ Proposing a multimodal ensemble framework for probabilistic vegetation loss prediction}. Existing approaches to wildfire impact prediction often rely on fewer than fifteen meteorological or static environmental variables. The proposed MVeLMA framework integrates heterogeneous data that captures both temporal weather patterns and static geographic variables into a multimodal architecture. It leverages temporal encoding, uncertainty-aware regression, and residual learning to provide vegetation loss forecasts with confidence intervals.

\item \textbf{Integrating a diverse set of feature variables}: MVeLMA combines nearly thirty-six dynamic and static variables, including meteorological trends, vegetation indices, and topographical features. This cross-modal design allows the model to capture granular regional differences and their influence on post-fire vegetation response.

\item \textbf{Extensively conducting empirical performance evaluations}: The proposed model was tested across more than fifty counties in a wildfire-prone state, demonstrating consistent improvement over baseline methods. It achieved an approximate 10\% increase in R\textsuperscript{2} and an 11\% reduction in NRMSE, highlighting the advantages of the multimodal architecture and uncertainty modeling.

\item \textbf{Generating post-wildfire risk maps}: This study produces county-level vegetation loss predictions with uncertainty estimates. These maps identify high-risk counties, enabling targeted post-fire recovery, resource allocation, and policy planning.

\end{itemize}

\section{Related Works}

The application of statistical modeling, machine learning (ML), and deep learning (DL) techniques in wildfire prediction and management has seen significant advancements. This section reviews recent literature, categorizing studies based on their methodological approaches.
\subsection{Statistical and ML Methods}
Traditional statistical and ML methods remain prevalent in wildfire prediction~\cite{jain_review_2020,bayat_comparison_2022,khanmohammadi_prediction_2022}. Joseph et al.~\cite{joseph_spatiotemporal_2019} apply Bayesian models to estimate extreme wildfire sizes by modeling upper quantiles from spatiotemporal predictors. Malik et al.~\cite{malik_data-driven_2021} utilize random forest models to predict wildfire risks by analyzing comprehensive datasets, including powerlines, terrain, and vegetation, to improve spatial and temporal accuracy. Gizatullin et al.~\cite{gizatullin_prediction_2022} construct linear models between fire risk and multiple factors. Castel-Clavera et al.~\cite{castel-clavera_disentangling_2022} design region-specific logistic regression model.   Zhang et al.~\cite{zhang_joint_2023} present a two-part model combining Integrated Nested Laplace Approximation (INLA) and Stochastic Partial Differential Equations (SPDE) to jointly model wildfire counts and burnt areas. Koh et al.~\cite{koh_spatiotemporal_2023} model wildfire ignitions using marked point processes. Illarionova et al.~\cite{illarionova_exploration_2025} combine random forest and XGBoost. Schmitt et al.~\cite{schmitt_ecosystem-based_2024} emphasize the role of ecological factors in fire susceptibility. Xu et al.~\cite{xu_adaptive_2025} employ spatiotemporal prediction and a non-parametric dynamic threshold. Ji et al.~\cite{ji_methods_2024} infer wildfire events from satellite images with a simple change-point analysis procedure.

\subsection{DL Methods}
Deep Learning Approaches for Wildfire Prediction
Several studies have leveraged deep learning architectures to enhance wildfire prediction accuracy~\cite{ferchichi_forecasting_2022}. Bhowmik et al.~\cite{bhowmik_multi-modal_2023} developed a multi-modal wildfire prediction and early-warning system using a U-Convolutional-LSTM neural network, achieving over 97\% accuracy by integrating environmental, meteorological, and geological data. Similarly, Burnt-Net~\cite{seydi_burnt-net_2022} utilized post-fire Sentinel-2 imagery with a deep learning morphological neural network for accurate burned area mapping. Deng et al.~\cite{deng_daily_2025} employed an attention mechanism and a graph convolution network for spatiotemporal daily wildfire risk prediction by mining global and local dependencies. CNNs are commonly used to infer fire impact from satellite images~\cite{priya_vegetation_2024}. Long Short-Term Memory (LSTM) networks are widely used for fire forecasting with time series data~\cite{reddy_prediction_2018,vasilakos_lstm-based_2022,lin_forest_2023}. Conv-LSTM-based methods~\cite{liu_forecasting_2023,kartal_next-level_2024,masrur_capturing_2024} are also popular methods in spatiotemporal data modeling. Yao et al.~\cite{yao_spatio-temporal_2022} design a 3D-CNN with LSTM. CNN-BiLSTM~\cite{marjani_cnn-bilstm_2024} performs daily spread prediction with remote sensing and weather data. Wang et al.~\cite{wang_prediction_2025} add attention mechanism (AM) on top of CNN-BiLSTM. Jadouli et al.~\cite{jadouli_enhancing_2024-1} introduce a modular transformer-based deep learning architecture with pretrained internal “world model” layers. Yu et al.~\cite{yu_predicting_2023} develop a spatiotemporal transformer model to predict PM2.5 levels in wildfire-prone regions. Rösch et al.~\cite{rosch_data-driven_2024} introduce a GNN-based framework to model wildfire spread across Europe using spatial node embedding and dynamic temporal graphs, allowing generalization across locations. Parellada et al.~\cite{parellada_i_calderer_predicting_2024} leverage a BiLSTM with a GNN. 

\subsection{Mixed Models and Data Sources}
Mixture of Models and Multi-Modal Data Integration
Integrating deep learning and machine learning models, as well as multi-modal spatiotemporal data~\cite{xu_spatio-temporal_2023} has been a focus in recent studies. For instance, Jadouli et al.~\cite{jadouli_enhancing_2024} propose a modular forecasting pipeline combining deep learning, ensemble methods, and transfer learning. Huot et al.~\cite{huot_next_2022} provide a dataset for wild fire spread prediction with remote sensing data while the dataset WildfireSpreadTS~\cite{gerard_wildfirespreadts_2023} includes multi-modal temporal sequences. Kabir~\cite{kabir_wildfire_2024} utilizes both ARIMA and LSTM. Jadouli et al.~\cite{jadouli_enhancing_2024-1} fuse CNNs, LSTMs, ensembles, and transfer learning into a unified wildfire forecasting framework.
Zhu et al.~\cite{zhu_assessing_2018} carefully process and identify important factors for defoliation forecasting. Dong et al.~\cite{dong_wildfire_2022} emphasize localized modeling using topographic and meteorological factors. Khanmohammadi~\cite{khanmohammadi_prediction_2022} incorporate vegetation type, slope, and meteorological data. Ghorbanzadeh et al.~\cite{ghorbanzadeh_spatial_2019} involve field GPS survey data. Zhang et al.~\cite{zhang_analyzing_2021} identify climate and human activity as key spatiotemporal drivers. Deng et al.~\cite{deng_wildfire_2023} incorporate rainfall, slope, vegetation, and historical ignitions. Michail~\cite{michail_seasonal_2024} performs seasonal fire prediction with vegetation dryness, past fires, and climate data. Parellada et al.~\cite{parellada_i_calderer_predicting_2024} introduce terrain and wind direction dynamics. Moghim et al.~\cite{moghim_wildfire_2024} highlight regional variability and the need for localized modeling. CREDS~\cite{john_resource-efficient_2025} is designed to handle wildfires with Unmanned Aerial Vehicles (UAVs). 

While prior research demonstrates advancements in spatiotemporal modeling, most approaches either focus on deterministic predictions or rely on isolated data sources. In contrast, MVeLMA integrates deep sequential encoding, uncertainty quantification, and multimodal inputs to produce region-specific and confidence-aware predictions of post-fire vegetation loss.

\begin{table*}
  \caption{Summary of data variables used in the study}
  \label{tab:weather-features}
  \begin{tabular}{p{3.2cm}p{1.5cm}p{3.8cm}p{1.8cm}p{1.4cm}p{1.8cm}}
    \toprule
    \textbf{Data Features} & \textbf{Units} & \textbf{Sources} & \textbf{Time Range} & \textbf{Usage} & \textbf{Importance (\%)} \\
    \midrule
    Average Temperature & °C & Univ. of Idaho (EPSCOR) \cite{kliskey_idaho_nodate} & 2018--2024 & Input & 5.30 \\
    Precipitation & mm & Univ. of Idaho (EPSCOR) & 2018--2024 & Input & 8.48 \\
    Relative Min Humidity & \% & Univ. of Idaho (EPSCOR) & 2018--2024 & Input & 14.75 \\
    Relative Max Humidity & \% & Univ. of Idaho (EPSCOR) & 2018--2024 & Input & 4.87 \\
    Solar Radiation & W/m\textsuperscript{2} & Univ. of Idaho (EPSCOR) & 2018--2024 & Input & 3.68 \\
    Minimum Temperature & °C & Univ. of Idaho (EPSCOR) & 2018--2024 & Input & 8.19 \\
    Maximum Temperature & °C & Univ. of Idaho (EPSCOR) & 2018--2024 & Input & 4.09 \\
    Vapor Pressure Deficit & kPa & Univ. of Idaho (EPSCOR) & 2018--2024 & Input & 3.88 \\
    Wind Speed & m/s & Univ. of Idaho (EPSCOR) & 2018--2024 & Input & 5.09 \\
    Elevation & m & USGS SRTMGL1 v003 \cite{farr_shuttle_2007} & 2018 & Input & 6.16 \\
    Land coverage & 0--1 ratio & MODIS MCD12Q1 v6.1 \cite{friedl_modisterraaqua_2022} & 2018&Input & 25.68 \\
    Fire Events & dates & MODIS C6.1 Fire Archive \cite{giglio_modis_nodate} & 2018--2024 & Input & 14.75 \\
    NDVI & scaled & MODIS MOD13Q1 v6.1 \cite{didan_modisterra_2021} & 2018--2024 & Target & -- \\
    \bottomrule
  \end{tabular}
\end{table*}

\begin{table}
  \caption{Data sources and respective spatial/temporal resolutions.}
  \label{tab:sources_resolution}
  \begin{tabular}{p{5cm} p{3cm}}
    \toprule
    \textbf{Source} & \textbf{Resolution} \\
    \midrule
    GRIDMET (Abatzoglou, 2013) & 4 km, daily \\
    SRTMGL1 v003 (Farr et al., 2007) & 30 m \\
    MODIS MCD12Q1 v6.1 (Friedl et al., 2019) & 500 m, annual \\
    MODIS MOD13Q1 v6.1 (Didan, 2015) & 250 m, 16-day composite \\
    MODIS MCD14ML (FIRMS) & 1 km, daily \\
    \bottomrule
  \end{tabular}
\end{table}

\section{Methodology}

\subsection{Preliminaries}
In this work, we aim to build a predictive model that not only focuses on predicting the binary event of wildfire but also its effects. To make the application more suitable for real-world use cases, we also focus on incorporating spatially coherent uncertainty estimation to guide policy-level interventions and optimize resource allocation during emergencies. Therefore, we use vegetation loss as an indicator of the impact of the fire and formulate it as a spatiotemporal multivariate time series forecasting problem.
The problem is formally defined as follows:
\newline
\textbf{Input}:
\begin{itemize}
\item A set of county-level wildfire events $F=\{f_1,f_2,f_3, ...,f_n\}$ where each $f_i$ represents a wildfire event within the respective county with spatial and temporal information.

\item A 30-day multivariate time series prior to each fire event $f_i$ represented by $X^{T}$. This 3-D array is of shape $N x T x W$ where $N$ represents the number of fire events, $T$ = 30, the 30-day sequence of meteorological history, and $W$ = 9, which represents the number of daily meteorological features.

\item Enriched features represented by $X^{S}$. This 2-D array is of shape $N x S$ where $N$ represents the number of fire events and $S$ = 27, which represents the number of land coverage features and fixed features.

\end{itemize}
Given X=\{$x_1,x_2,..x_N$\}, Let $\mathbf{x}_i^T \in \mathbb{R}^{T \times W}$ denote the multivariate time series for $T=30$ days and $W$ meteorological features, and $\mathbf{x}_i^S \in \mathbb{R}^{S}$ represent $S$ enriched spatial features (e.g., land cover, elevation). The complete input is given by:
\[
\mathbf{x}_i = \left( \mathbf{x}_i^T, \mathbf{x}_i^S \right)
\]

\textbf{Target Variable:} Vegetation loss is measured using NDVI values.
NDVI loss values for each fire event $f\in F$, measured as $y$ = $ V_{abf} - V_{maf}$ where $ V_{abf} $ refers to the mean NDVI for 30 days before fire event $ f $ and $V_{maf} $ refers to the minimum NDVI within the 30 days after the fire event. \\
\textbf{Problem formulation:} We formulate post-fire vegetation loss prediction as a supervised regression problem over a multiscale feature space that captures both temporal dynamics and spatial heterogeneity. Let $\mathcal{D} = \{(\mathbf{x}_i, y_i)\}_{i=1}^{N}$ denote our dataset, where each $\mathbf{x}_i$ encodes spatial location, 30-day pre-fire meteorological and topographical time series, and enriched county-level geophysical descriptors for wildfire event $f_i$, and $y_i \in \mathbb{R}$ denotes the observed NDVI loss.

\textbf{Objective}:
\begin{itemize}
\item Maximize the accuracy and precision of the stacked time series model for estimating post-fire vegetation loss
\item Calculate and aggregate coherent uncertainty measures at the county-level to support identification of at-risk and vulnerable regions.
\end{itemize}

\subsection{Model development} \label{modeldevelopment}
We adopt a multimodal data integration pipeline that couples with a stacked modeling approach in which model integration is based on the modality of the data and the latent feature space. This stacked architecture leverages the strength of each component for the various data modalities (i.e., temporal, spatial, and static).
 
The overall workflow is illustrated in Figure \ref{fig:model_pipeline}. The model is a stacked model comprising an encoding layer, GPR, and a Random Forest (RF) regression layer.

\subsubsection{\textbf{Integrating Temporal Weather Data through Recurrent Sequential Modeling}} 
Long Short-Term Memory (LSTM) is a type of Recurrent Neural Network (RNN), known for its efficacy in capturing long-term temporal dependencies \cite{hochreiter_long_1997}. Like LSTM, BiLSTM is effective at modeling long-term dependencies in time series data \cite{huang_bidirectional_2015}, making it effective for incorporating 30-day weather sequences preceding fire events. However, BiLSTM extends the traditional LSTM architecture, allowing the model to capture temporal dependencies by processing input sequences in both forward and backward directions. This allows the model to identify vegetation patterns that may not be strictly causal or unidirectional. This bidirectional context improves the model's ability to capture the temporal relationship between fire and climate features, considering the influence of both preceding and succeeding weather patterns on fire conditions.

This layer processes a 3-D tensor input with dimensions N × T × W, where N is the number of fire events in the training set, T is the 30-day sequence length, and W is the number of weather features (10) at each time step. 
To extract temporal signatures from $\mathbf{x}_i^T$, the Bidirectional LSTM (BiLSTM) network computes a sequence of hidden states $\{\mathbf{h}_{it}\}_{t=1}^{T}$ in both forward and backward directions. The attention mechanism helps the model focus on the most important features in the data and the most informative time steps. Therefore, a context vector $\mathbf{c}_i \in \mathbb{R}^{d_h}$ is computed via attention weights $\alpha_{it}$:
\begin{align}
\mathbf{c}_i = \sum_{t=1}^{T} \alpha_{it} \mathbf{h}_{it}, \quad \alpha_{it} = \frac{\exp(e_{it})}{\sum_{k=1}^{T} \exp(e_{ik})}, \quad e_{it} = \text{score}(\mathbf{h}_{it})    
\end{align}

This context vector is then projected to a latent representation $\mathbf{z}_i \in \mathbb{R}^{D}$ via a dense layer:
\begin{align}
\mathbf{z}_i = \phi(\mathbf{W} \mathbf{c}_i + \mathbf{b})
\end{align}
where $D=20$ is the latent dimension, and $\phi$ is a non-linear activation. This latent vector is an encoding of the weather sequence, structured to capture temporal trends involved in vegetation loss.
\subsubsection{\textbf{Modeling Spatial Dependency through Multivariate Gaussian Process}} 
GPR is a type of non-parametric probabilistic method for regression and time-series forecasting problems due to its flexibility in quantifying predictive uncertainty \cite{guo_time_2012}. It defines prior probability distributions over latent functions and uses Bayes' Theorem to calculate the posterior distribution given observed data. The prior distribution is characterized by the regression mean function and a covariance matrix.  The covariance matrix is developed by a kernel function that encodes information about the shape, periodicity, and structure that the function is expected to have. For our pipeline, we take the output of the previous module, i.e., the latent vector $\mathbf{z}_i$, and concatenate it with enriched features $\mathbf{x}_i^S$ to form $\tilde{\mathbf{x}}_i = [\mathbf{z}_i ; \mathbf{x}_i^S] \in \mathbb{R}^{D + S}$.
The GPR models ground truth vegetation loss values ($y_i$) as a distribution over functions in the following way:  
\begin{align}
f(\cdot) \sim \mathcal{GP}(m(\cdot), k(\cdot,\cdot))     
\end{align}
where $m(\cdot)$ is the mean function (set to constant) and $k(\cdot,\cdot)$ is the kernel function.
In our experiments, we explore multiple kernel choices and their combination:
\begin{itemize}
 \item \textbf{RBF kernel (baseline)}:
    \begin{align}
    k_{\text{RBF}}(\mathbf{\tilde{x}}_i, \mathbf{\tilde{x}}_j) = \sigma^2 \exp\left( -\frac{\|\mathbf{\tilde{x}}_i - \mathbf{\tilde{x}}_j\|^2}{2\ell^2} \right)    
    \end{align}
    The RBF kernel is widely used in prediction and regression problems because it effectively maps the data onto an infinite-dimensional feature space, making it flexible. This distance function-based kernel is a more compactly supported kernel function, thereby reducing the computational complexity of the training process \cite{sohrabi_predicting_2023}.
    \item \textbf{Matérn kernel}:
    \begin{align}
  k_{\text{Mat}}(\mathbf{\tilde{x}}_i, \mathbf{\tilde{x}}_j) = \sigma^2 \cdot \frac{2^{1-\nu}}{\Gamma(\nu)} \left( \sqrt{2\nu} \frac{\|\mathbf{\tilde{x}}_i - \mathbf{\tilde{x}}_j\|}{\ell} \right)^\nu K_\nu\left( \sqrt{2\nu} \frac{\|\mathbf{\tilde{x}}_i - \mathbf{\tilde{x}}_j\|}{\ell} \right)
    \end{align}
    with $\nu = 2.5$, allowing for a balance between smoothness and flexibility.
    
    \item \textbf{Composite kernel (Matérn + Periodic)}:
    \begin{align}
    k_{\text{comp}} = k_{\text{Mat}} + k_{\text{Per}}, \quad
    k_{\text{Per}}(\mathbf{\tilde{x}}_i, \mathbf{\tilde{x}}_j) = \exp\left( -\frac{2\sin^2(\pi \|\mathbf{\tilde{x}}_i - \mathbf{\tilde{x}}_j\| / p)}{\ell^2} \right)    
    \end{align}
    
The composite kernel was chosen to augment the Matérn kernel with the periodic kernel's ability to explicitly model data that exhibits seasonality.
\end{itemize}
Primarily, the GPR module produces two outputs for each input $\tilde{\mathbf{x}}_i$. First, the posterior predictive mean $\mu_i$, and second, the predictive variance $\sigma_i^{2}$. The mean $\mu_i$ represents the model's expected value of NDVI loss for the fire event $i$, which incorporates both temporal dynamics (via $\mathbf{z}_i$) and spatial features (via $\mathbf{x}_i^S$). The predictive variance represents epistemic uncertainty. These outputs are computed using standard Gaussian Process posterior inference equations conditioned on the training data $\mathbf{X}$ for a test location $\tilde{\mathbf{x}}^*$:
\begin{align}
\mu^* &= k(\tilde{\mathbf{x}}^*, \mathbf{X})^\top \left[ K + \sigma_n^2 \mathbf{I} \right]^{-1} \mathbf{y} \\
\sigma^{*2} &= k(\tilde{\mathbf{x}}^*, \tilde{\mathbf{x}}^*) - k(\tilde{\mathbf{x}}^*, \mathbf{X})^\top \left[ K + \sigma_n^2 \mathbf{I} \right]^{-1} k(\tilde{\mathbf{x}}^*, \mathbf{X})
\end{align}
where, $\mu^*$ is the Predictive mean of NDVI loss for test input $\tilde{\mathbf{x}}^*$; $\sigma^{*2}$ denotes the predictive variance (uncertainty) for test input $\tilde{\mathbf{x}}_i$; $k(\tilde{\mathbf{x}}^*, \mathbf{X})$ is the Kernel vector between test input $\tilde{\mathbf{x}}_i$ and all training inputs $\mathbf{X}$; $k(\tilde{\mathbf{x}}^*, \tilde{\mathbf{x}}^*)$ is the Kernel evaluated at the test input with itself (i.e., prior variance at $\tilde{\mathbf{x}}^*$); $K = k(\mathbf{X}, \mathbf{X})$ represents the covariance Kernel matrix (Gram matrix) of all training inputs; $\sigma_n^2$: Noise variance hyperparameter, learned during training; $\mathbf{I}$: Identity matrix of shape $N \times N$ and $\mathbf{y}$: Vector of ground truth NDVI loss values for the training data.

\subsubsection{\textbf{Random Forest Regression for Residual Connection and Robustness:}}

This module addresses the limitation of deep neural networks’ deterministic predictions and limited capacity of a probabilistic learner to model high-frequency interaction in low-data regions. We concatenate the posterior mean from the GPR with the  latent vector derived from the Bi-LSTM layer to construct the input for an RF module. 
\begin{align}
    \mathbf{x}_i^{\text{RF}} = [\mathbf{z}_i ; \mu^*_i] \in \mathbb{R}^{D+1}
\end{align}
RF is an ensemble model consisting of multiple decision trees that learn the patterns from the combined input and output a final vegetation loss prediction for each test sample by calculating the collective average of the decision trees. The incorporation of RF enables the model to be generalizable to spatially heterogeneous regions where Gaussian Process may underperform. Furthermore, it makes the model robust to overfitting and capable of performing implicit feature selection. Additionally, RF has lower sensitivity to hyperparameters, requiring minimal fine-tuning to achieve strong performance compared to architectures such as multi-layer perceptrons (MLP) and support vector machines (SVM) \cite{you_comparison_2017}. Lastly, it serves as a computationally efficient refinement for final prediction by training it as a residual learner.
\begin{align}
\hat{y}_i = \text{RF}(\mathbf{x}_i^{\text{RF}})
\end{align}

The final vegetation loss predictions from the RF layer are validated against the observed vegetation loss. The results of the validation will indicate how accurately the hybrid stacked model performs. 

\begin{algorithm}
\caption{MVeLMA Joint Optimization Model}
\label{alg:mvelma}
\begin{algorithmic}[1]
\For{each fire event $f_i$}
    \State Get temporal sequence $\mathbf{x}_i^T$, enriched features $\mathbf{x}_i^S$, and latent features vector $\mathbf{z}_i$
    \State Compute hidden states: BiLSTM$(\mathbf{x}_i^T) \rightarrow \{\mathbf{h}_{it}\}_{t=1}^{T}$
        \State Compute attention weights $\quad e_{it} = \text{Linear}(\mathbf{h}_{it})$
        \State Normalize attention weights: softmax($e_{it}$) $\rightarrow \alpha_{it}$
        \State Compute context vector $\mathbf{c}_i = \sum_{t=1}^{T} \alpha_{it} *\mathbf{h}_{it}$ \Comment{ Weighted sum of hidden states}
        \State Encode latent vector: $\text{Linear}(\mathbf{c}_i) \in \mathbb{R}^D \rightarrow \mathbf{z}_i $
    \State Train GPR on $\mathbf{z}_i $
    \For{epoch $e < 500 $}
        \State Compute GP posterior mean $\mu_i$
        \State Backpropagate and optimize: $\mathcal{L}_e = -\text{MLL}(\mu_i, y_i)$
        \If{early stopping criteria is met} \Comment{ uses patience and minimum loss as criteria}
            \State \textbf{break}
        \EndIf
    \EndFor
    \State Predict GPR posterior mean: $\mu_i$ and variance $\sigma^2_i$
    \State Stack features: ${\mathbf{x}}_i^{RF} \leftarrow [\mathbf{z}_i, \mathbf{x}_i^S, \mu_i]$
    \For{each RF iteration} \Comment{Train RF to correct GP prediction errors}
        \State Train RF on ${\mathbf{x}}_i \rightarrow$ residual
        \State Predict residual 
        \State Final prediction: $\hat{y}_i = \text{RF}(\mathbf{x}_i^{RF})$ 
    \EndFor
    \State Evaluate: MAE, R\textsuperscript{2}, MAPE, NRMSE
\EndFor
\end{algorithmic}
\end{algorithm}

\subsubsection{\textbf{Joint Optimization Strategy}} \label{joint_optimization}

While each component in our architecture incorporates different modalities (temporal sequences, spatial metadata, etc.), we ensure coherent integration of these modules through a sequential stacking strategy. Algorithm~\ref{alg:mvelma} presents pseudocode summarizing the three phases of the training pipeline.

\paragraph{Loss Function:}
We optimize the GPR layer on the latent features produced by the BiLSTM encoder. The loss function is modeled by the negative log marginal likelihood (MLL) of the GPR posterior mean.
\begin{align}
L_{GPR_by_BiLSTM}=-\frac{1}{2}y_T(\left[ K + \sigma_n^2 \mathbf{I} \right]^{-1} \mathbf{y})+\frac{1}{2} \log|K+ \sigma_n^2 I|+\frac{n}{2} \log(2\pi)
\end{align}
This loss function is minimized when the BiLSTM is trained with the GPR, as its parameters are updated through backpropagation based on the GPR’s marginal likelihood loss using an optimizer. The GPR posterior mean, BiLSTM-encoded latent vectors, and enriched features are concatenated and passed to the RF model. This model is trained to minimize the residual error between the GPR posterior mean and the ground truth value, using a mean squared error loss function.
\begin{align}
L_{RF}=\frac{1}{n} \sum_{i=1}^{n}(\hat{y}_i-{y}_i)^2
\end{align}

\subsection{Experiments}
In this section, we present the experimentation process we have conducted using wildfires in the state of California to illustrate the accuracy of our algorithms and compare qualitatively the vegetation loss prediction using a curated set of candidate models. We have implemented the stacked deep learning model using multivariate time series, as described in Section \ref{modeldevelopment}, and provide a comprehensive analysis of the performance of the model along with the role of each layer.

\begin{table}[t]
\centering
\small
\caption{Enriched features used in the RF model alongside learned latent representations.}
\label{tab:enriched-features}
\begin{tabular}{>{\raggedright\arraybackslash}p{2.2cm} p{5.8cm}}
\toprule
\textbf{Feature Name} & \textbf{Description} \\
\midrule
fire\_duration\_days & Length the fire lasted in days \\
fire\_month & Month the fire started \\
fire\_dayofyear & Day of the year the fire started \\
ndvi\_7d\_slope & NDVI slope over the 7 days before the fire \\
ndvi\_7d\_std & NDVI standard deviation over the 7 days before the fire \\
ndvi\_14d\_slope & NDVI slope over the 14 days before the fire \\
ndvi\_14d\_std & NDVI standard deviation over the 14 days before the fire \\
ndvi\_before\_slope & NDVI slope over the 30 days before the fire \\
ndvi\_before\_std & NDVI standard deviation over the 30 days before the fire \\
elevation & Average elevation of the county where the fire occurred \\
LC\_00 \cite{devadiga_landwebmodapseosdisnasagovtsdatac6_schemehtml_nodate} & Water \\
LC\_01 & Evergreen needleleaf forest \\
LC\_02 & Evergreen broadleaf forest \\
LC\_03 & Deciduous needleleaf forest \\
LC\_04 & Deciduous broadleaf forest \\
LC\_05 & Mixed forests \\
LC\_06 & Closed shrublands \\
LC\_07 & Open shrublands \\
LC\_08 & Woody savannas \\
LC\_09 & Savannas \\
LC\_10 & Grasslands \\
LC\_11 & Permanent wetlands \\
LC\_12 & Croplands \\
LC\_13 & Urban and built-up \\
LC\_14 & Cropland/natural vegetation mosaic \\
LC\_15 & Snow and ice \\
LC\_16 & Barren or sparsely vegetated \\
\bottomrule
\end{tabular}
\end{table}

\subsubsection{Data collection and processing}
California was selected as the study area due to its recurrent, large-scale wildfires and the availability of high-resolution environmental and remote sensing data, making it a representative and data-rich region for modeling vegetation loss. Effective modeling of vegetation loss requires integrating core fire event data with relevant meteorological and topographical features that influence a wildfire's impact on vegetation. This study incorporates diverse datasets covering wildfire behavior and environmental features using Google Earth Engine (GEE) scripts. The components are described below, with Table~\ref{tab:weather-features} summarizing key features and Table~\ref{tab:sources_resolution} detailing their spatial and temporal resolutions. Feature importance analysis was conducted using an RF model trained on aggregate feature representations to assess the contribution of individual meteorological and topographical features to model performance. Specifically, importance was calculated using Gini-based measures \cite{druckman_measuring_2008}. As shown in Table \ref{tab:weather-features}, aggregate land coverage features and historical fire events greatly influence predictors (23.68\% and 14.75\% respectively), while meteorological features show varying contributions to model performance.
\paragraph{NDVI}
Vegetation productivity, measured by the Normalized Difference Vegetation Index (NDVI), is derived from remote sensing data that assesses vegetation density and health based on light reflectance. NDVI values range from -1 to 1, where higher values indicate dense and healthy vegetation, and lower values correspond to sparse, stressed, or barren surfaces \cite{ozyavuz_determination_2015}.
NDVI reflects vegetation dynamics influenced by seasonal patterns, and its values can fluctuate depending on the time of year, regardless of fire activity. Additionally, the 16-day sampling interval of the satellite introduces temporal misalignment with fire events and daily weather data. To address both the coarse temporal resolution and the underlying seasonality, NDVI values from the 30 days before and after each fire were aggregated using the mean and minimum, respectively, without interpolation.

\paragraph{Fire events}
Fire data derived from satellite observations were filtered to retain only high-confidence detections (i.e., confidence $\geq$ 60), ensuring the reliability of identified fire pixels. These detections were then mapped to historical wildfire records in California using the California Open Data Portal \cite{california_department_of_forestry_and_fire_protection_cal_fire_and_its_fire_and_resource_assessment_program_frap_california_nodate}. For each fire event, the acquisition date and geographic coordinates were extracted and mapped to their corresponding counties. To enhance forecasting performance, additional temporal features were derived, as summarized in Table \ref{tab:enriched-features}.

\paragraph{Meteorological data}
Nine climate and weather variables were selected, each available as daily measurements at the county level. To handle partially missing values within the aggregation window (i.e., 7 days before to 14 days after the fire), the mean of each weather variable was computed using the available daily values for the respective fire event and county. Fire events with completely missing weather data across any one of the selected variables were excluded from the final dataset.

\paragraph{Topographical data}
County-level mean elevation and seventeen land cover fractions for the year 2018 were incorporated as enriched static features. Missing elevation values were imputed using the overall mean elevation across all counties. For land cover fractions, missing values were interpreted as the absence of the corresponding class and, therefore, replaced with zeros.
\begin{table}[t]
\begin{minipage}{\columnwidth}
\centering
\small
\caption{Comparison of MVeLMA with Baseline Methods}
\label{tab:baseline-onecol}
\begin{tabular}{>{\raggedright\arraybackslash}p{2.8cm} p{0.7cm} p{0.7cm} p{0.7cm} p{0.7cm} p{0.7cm}}
\toprule
\textbf{Model} & \textbf{MAE} & \textbf{R$^2$} & \textbf{MAPE (\%)} & \textbf{NRMSE} \\
\midrule
\textbf{MVeLMA} & \textbf{0.0109} & \textbf{0.7232} & \textbf{4.77} & \textbf{0.5261} \\
FFNN & 0.0126 & 0.6175 & 5.45 & 0.6185 \\
UFSTP  & 0.0151 &  0.4815 & 9.77 & 0.7201 \\
GRU & 0.0133 &  0.4850 & 5.15 & 0.7176 \\
1D-CNN & 0.0138 & 0.5382 & 6.87 & 0.6795 \\
LSTM-MC & 0.0159 & 0.4410 & 10.77 & 0.7477 \\
\bottomrule
\end{tabular}
\end{minipage}
\end{table}

\subsubsection{Baselines}
The performance of MVeLMA in comparison with both SOTA deterministic and stochastic models, which adopt various strategies for handling spatiotemporal data and integrating feature sets, is summarized in Table \ref{tab:baseline-onecol}. The details about the baselines are listed as follows:
\begin{itemize}
\item{Deterministic:}
    \begin{itemize}
        \item \emph{Dokhan et al.} 
        \cite{dokhan_neural_2024}: A feed-forward neural network model (FFNN) comprising fully connected hidden layers with nonlinear activation functions that was developed to forecast the burned area of forest fires using meteorological and spatiotemporal inputs.
        
         \item \emph{Gated Recurrent Units (GRU)} \cite{perumal_comparison_2020}: A type of RNN composed of gated recurrent units, well-known for capturing dynamic temporal dependencies.

        \item \emph{ One-dimensional CNN (1D-CNN)} \cite{schmieg_time_2024}: A neural network model that leverages temporal convolutional layers to identify temporal patterns in multivariate time series and extract spatial features between time steps, which has been applied extensively for time series forecasting. 
    \end{itemize}
\item{Stochastic:}
    \begin{itemize}
         \item \emph{Urban Fire Spatial–Temporal Prediction (UFSTP)} \cite{xiang_urban_2025}: A SOTA spatiotemporal and multi-source data enhanced predictive model. UFSTP uses an multi-layer perceptron (MLP) to encode enriched spatial features and Gated Recurrent Units (GRU) to encode dynamic temporal sequences.  It integrates a spatiotemporal propagation mechanism to model interactions across regions over time, concatenating both enriched and temporal representations to forecast fire occurrence.

        \item \emph{Long Short-Term Memory-Monte Carlo (LSTM-MC)} \cite{sattari_probabilistic_2025}: A SOTA probabilistic framework for one-step-ahead daily streamflow forecasting that uses a Monte Carlo-based deep learning model based on LSTM to account for model uncertainty.
        \end{itemize}
\end{itemize}
\subsubsection{Model Configuration}
The proposed model was implemented in PyTorch and trained on a CPU. The first component is a BiLSTM that encodes 30-day weather sequences prior to each fire event. The hidden size is set to 64 ($d_h=64$), resulting in a 128-dimensional output from the bidirectional LSTM. An attention mechanism applies a linear layer to each time step, followed by a softmax operation across time to compute attention weights. These weights are used to obtain a weighted sum of the BiLSTM outputs, which is passed through a fully connected layer to produce a 20-dimensional latent representation ($l=20$).

The second component, the GPR layer, is implemented using GPyTorch \cite{gardner_gpytorch_2021}. To model the target variable, vegetation loss, we used a Matérn kernel with a smoothness parameter of $\nu=2.5$, chosen for its flexibility and ability to capture spatial and temporal variability. The kernel is wrapped in a ScaleKernel to capture the overall magnitude of variation in the target function. The GPR model was trained using the Adam optimizer \cite{kingma_adam_2017} for up to 80 epochs with a learning rate of 0.1, minimizing the exact negative MLL to optimize the kernel parameters and the posterior mean function.

For each fire event, the input to train the RF was assembled by concatenating three features to form a 48-dimensional feature vector as follows: First, a 20-dimensional latent vector produced by the BiLSTM with an attention layer; second, a 27-dimensional vector of enriched covariates of shape $(F, 27)$, where $F$ represents the total number of fire events in the dataset shown in Table \ref{tab:enriched-features}; and finally, scalar value representing the posterior mean prediction derived from the GPR layer.
The RF was fine-tuned to balance model complexity and ensure stable predictions, using 500 trees. Parameters were selected to prevent under- and overfitting while also capturing dependencies between latent and enriched features. 
\subsubsection{Model Performance Comparison}
We compare our model with the five aforementioned SOTA models and traditional multivariate time series models typically applied to natural disaster forecasting (i.e., FFNN, UFSTP, GRU, 1D-CNN, LSTM-MC). Model performance was evaluated using four metrics: mean absolute error (MAE), coefficient of determination (R\textsuperscript{2}), mean absolute percentage error (MAPE), and normalized root mean squared error (NRMSE).
Based on the results, we have described several observations as follows.

First, as summarized in Table \ref{tab:baseline-onecol}, MVeLMA achieves a consistent lead in all four metrics. The improvement can largely be attributed to the model’s ability to integrate multimodal feature sources and capture their dependencies through distinct modeling components. Conversely, most existing time-series–based approaches for vegetation estimation rely on a limited set of features, typically 10 to 13 variables including spectral bands, meteorological indicators, and vegetation indices, and rarely explore how different model types align with different feature modalities \cite{h_nguyen_landsat_2020}.

The proposed model outperforms existing approaches by addressing key limitations in both probabilistic and deterministic baselines. Probabilistic methods often struggle to generalize in regions with limited data availability. Deterministic models, while effective in capturing temporal patterns from meteorological and topographical variables, fail to model spatial dependencies and lack the ability to quantify predictive uncertainty.
\subsubsection{Scalability Studies}
\begin{table}[t]
  \caption{Comparison of time cost (in seconds) of MVeLMA and baseline models}
  \label{tab:timescomparison}
  \centering
  \begin{tabularx}{\columnwidth}{>{\raggedright\arraybackslash}X
      S[table-format=1.3]
      S[table-format=2.3]
      S[table-format=1.2e-2]}
    \toprule
    {Model} & {Avg./epoch (s)} & {Training (s)} & {Eval (s)} \\
    \midrule
    \textbf{\makecell[l]{MVeLMA\\(GPR Layer only)}}
      & \textbf{\num{0.29}}  & \textbf{\num{23.23}} & \textbf{\num{2.88e-05}} \\
    FFNN
      & \num{0.002} & \num{1.260}  & \num{1e-03} \\
    UFSTP
      & \num{0.002} & \num{0.030} & \num{5e-03} \\
    GRU
      & \num{0.610} & \num{84.33} & \num{9e-03} \\
    1D-CNN
      & \num{0.250} & \num{23.37} & \num{5e-03} \\
    LSTM-MC
      & \num{0.650} & \num{83.58} & \num{9e-03} \\
    \bottomrule
  \end{tabularx}

\end{table}
Beyond the performance of the predictions, the computational efficiency is a vital component of real-world use cases. In this stage, we compare the runtime of each of the three layers, the time stratified by kernel choice, and the train and test times of MVeLMA with five baselines. As shown in Table~\ref{tab:timescomparison}, MVeLMA's training time per epoch is comparable to that of the candidate models. However, its inference time is over 90\% faster than the fastest among them (i.e., the FFNN), demonstrating its efficiency at prediction time.

\subsubsection{Ablation Studies and Cross-regional Validation Study}
In this subsection, we present ablated studies to evaluate the effectiveness of each component of the proposed model. We design six variants of MVeLMA and compare them with the complete version of MVeLMA. The differences between these seven models are described as follows:
\begin{itemize}
    \item \emph{w/o BiLSTM}: removing the BiLSTM component i.e., no longer leveraging the pre- and post-fire temporal sequence
    \item \emph{w/o GPR}: removing the GPR component
    \item \emph{w/o RF}: removing the RF component i.e., no longer probabilistically modeling how latent features relate to vegetation loss 
    \item \emph{w/o BiLSTM and GPR}: removing both BiLSTM and GPR i.e., using only RF prediction 
    \item \emph{w/o BiLSTM and RF}: removing BiLSTM and RF i.e., using only GPR 
    \item \emph{w/o GPR and RF}: removing GPR and RF i.e., using only BiLSTM 
    \item \emph{MVeLMA}: the final version of the proposed model
    
\end{itemize}
\begin{table}[t]
\centering
\small
\caption{Ablation Study of Model Components on Vegetation Loss Prediction}
\begin{tabular}{p{2.8cm} p{0.7cm} p{0.7cm} p{0.7cm} p{0.7cm} p{0.7cm}}
\toprule
\textbf{Model Variant} & \textbf{MAE} & \textbf{R$^2$} & \textbf{MAPE (\%)} & \textbf{NRMSE} \\
\midrule
\textbf{MVeLMA} & \textbf{0.0109} & \textbf{0.7232} & \textbf{4.77} & \textbf{0.5261} \\
\quad w/o BiLSTM                        & 0.0111 & 0.6567 & 4.8244 & 0.5859 \\
\quad w/o GPR                          & 0.0113 & 0.6832 & 5.1196 & 0.5629 \\
\quad w/o RF                           & 0.0121 & 0.6250 & 6.4640 & 0.6124 \\
\quad w/o BiLSTM and GPR               & 0.0116 & 0.6677 & 5.2043 & 0.5764 \\
\quad w/o BiLSTM and RF                & 0.0127 & 0.6118 & 5.8763 & 0.6230 \\
\quad w/o GPR and RF                   & 0.0137  & 0.5446 & 5.5768 & 0.6748 \\
\textbf{MVeLMA (Oregon Validation} & \textbf{0.0247} & \textbf{0.7390} & \textbf{1.51} & \textbf{0.5109} \\
\bottomrule
\end{tabular}
\label{tab:ablation}
\end{table}

\begin{figure}[t]
  \centering
  \includegraphics[width=\linewidth]{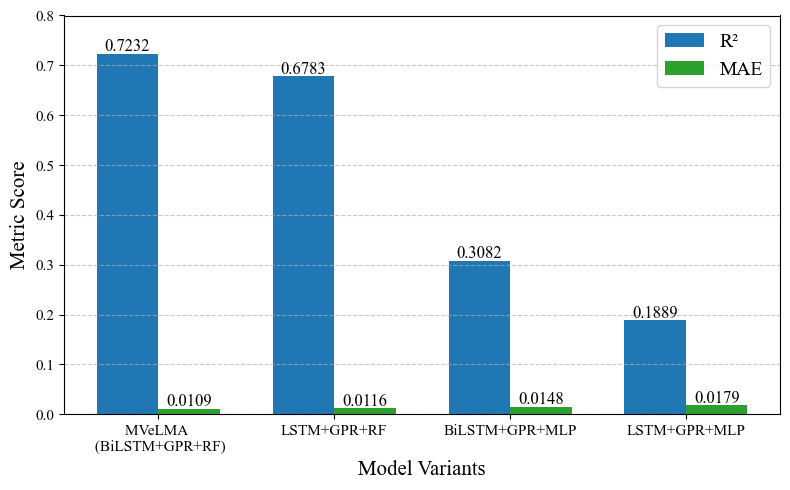}
  \Description{Bar chart comparing MVeLMA, LSTM+GPR+RF, BiLSTM+GPR+MLP, and LSTM+GPR+MLP models across R-squared and MAE metrics. The plot shows BiLSTM outperforming unidirectional LSTM, and RF outperforming MLP.}
  \caption{Performance analysis across different architectural variations across R\textsuperscript{2} and MAE metrics. Results highlight the effectiveness of BiLSTM over unidirectional LSTM for capturing temporal fire event and climate dependencies, and RF over MLP for ensemble prediction.}
  \label{covariates}
  \vspace{-10pt}
\end{figure}
\noindent Table \ref{tab:ablation} summarizes the results. We delineate our observations as follows: (i) Replacing the BiLSTM with a 30-day mean-aggregated meteorological sequence reduces complexity but limits the model’s ability to capture temporal dynamics such as seasonal shifts or weather changes around fire events. This simplification leads to a modest performance drop (MAPE: 4.82\%, R\textsuperscript{2}: 0.6567 compared to 4.77\% and 0.7232, respectively). (ii) By removing the GPR layer, the model fails to capture uncertainty, reducing its adaptability to less frequently observed or spatially distinct conditions. Without the GPR posterior mean as an input to RF, the overall feature richness is also reduced, further lowering performance. (iii) Excluding the RF layer removes both the final refinement step and the contribution of enriched static features. This variant exhibits a significant performance decline (MAPE: 6.46\%, R\textsuperscript{2}: 0.6250), highlighting the importance of integrating contextual information through post-GPR correction. (iv) Finally, a purely deterministic RF model, using only aggregated meteorological and enriched features, performs better than the ablated variants but lacks uncertainty estimation and fails to capture temporal dependencies. While reasonably competitive (MAPE: 5.26\%, R\textsuperscript{2}: 0.6677), it underperforms compared with the MVeLMA model, reiterating the advantage of a stacked ensemble model.

\begin{figure*}[t]
    \centering
    \includegraphics[width=\textwidth]{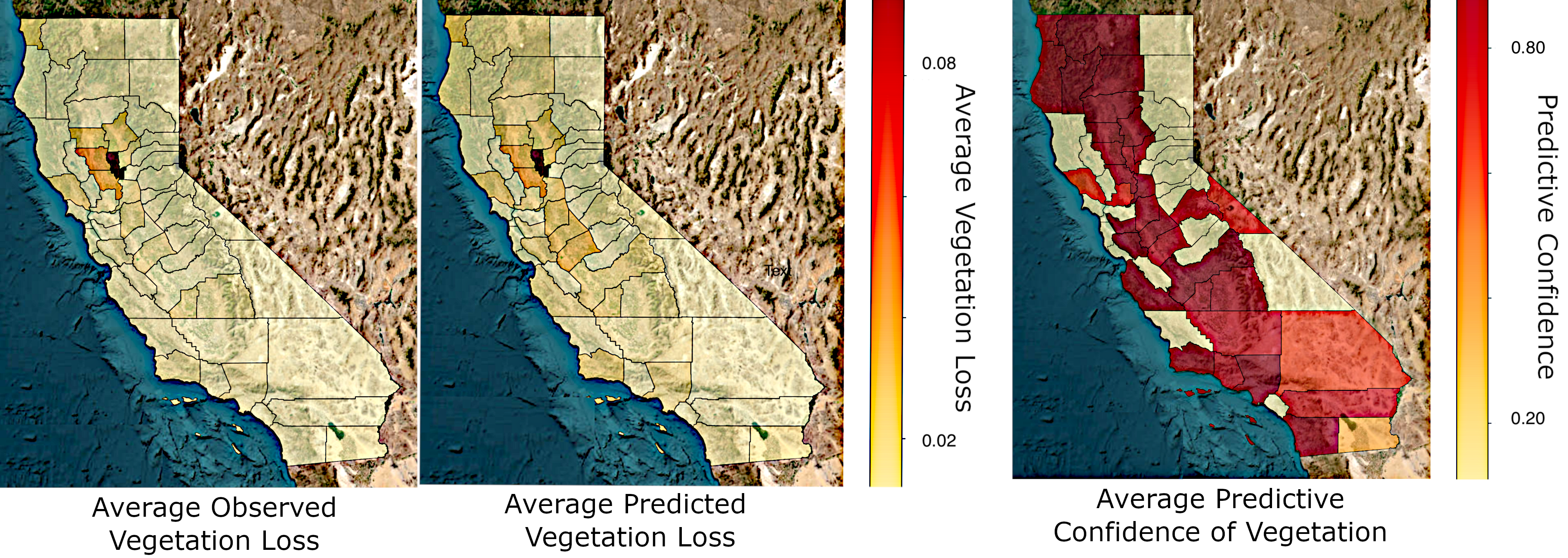}
    \caption{This figure displays the spatial distribution of MVeLMA's observed, predicted, and predictive confidence estimates across all California counties, where darker colors represent higher vegetation loss and predictive confidences respectively.}
    \Description{Visual geographic map of California counties showing observed vegetation loss, predicted vegetation loss, and confidence estimates. Darker colors indicate higher values.}
    \label{fig:California}
\end{figure*}

The predictive accuracy of MVeLMA was compared against architectural variants, specifically using unidirectional LSTM and MLP instead of BiLSTM and RF, respectively. As shown in Figure \ref{covariates}, replacing BiLSTM with LSTM resulted in a performance decrease in R\textsuperscript{2} and MAE scores, highlighting the effectiveness of the bidirectional approach in capturing temporal dependencies between fire events and climate patterns. Additionally, using MLP instead of RF significantly lowered performance, with prediction accuracy decreasing by nearly 50\%. This underscores the robustness of RF and its reduced sensitivity to hyperparameter tuning.

To evaluate the model's cross-regional adaptability, MVeLMA was tested on vegetation loss in Oregon, a state with a wetter, more temperate climate and different elevation and vegetation patterns compared to California \cite{usda_climate_hubs_climate_nodate}. The Oregon validation used 64 wildfire events from May to July 2023, with metrics shown in Table \ref{tab:ablation}. The R\textsuperscript{2} value for this region was comparable to that of California at 0.7390, suggesting MVeLMA's architecture effectively captures the relationship between topographical and climate features and wildfire events. Specifically, the model achieved lower MAPE (1.51 vs. 4.77), indicating more consistent predictions across the different regional fire patterns. However, the higher MAE (0.0247 vs. 0.0109) suggests that while the model captures the relationship between the covariates and wildfire events, additional fine-tuning would likely be required to better account for region-specific landscape variances.

Additionally, to validate temporal generalization, MVeLMA was tested on Oregon data from an earlier seasonal period (May to July) compared to the California training period (July to September). This tests the model's adaptability to not only geographical variations but also seasonal variations in fire conditions, as fire seasonality patterns vary between Oregon's wetter climate and California's drier Mediterranean climate.

\subsection{Case Study}
To evaluate model performance and interpretability, we conducted a spatial and temporal analysis of post-fire vegetation loss across selected California counties using observed post-fire vegetation loss (OPFVL), predicted post-fire vegetation loss (PPVL), and predictive confidence (APC).

Across the test counties, OPFVL ranged from {0.0006} (Alameda County) to {0.0873} (Colusa County), with higher values concentrated in counties experiencing larger, more severe fires. The model’s PPVL ranged from {0.0092} to {0.0684}, while APC remained consistently high, between {0.87} and {0.98}. Figure \ref{fig:California} illustrates this spatial distribution, showing that counties with larger, more frequent fire events exhibited higher predictive confidence, likely due to more training data.
\begin{figure*}[h!]
    \centering
    \includegraphics[width=\textwidth]{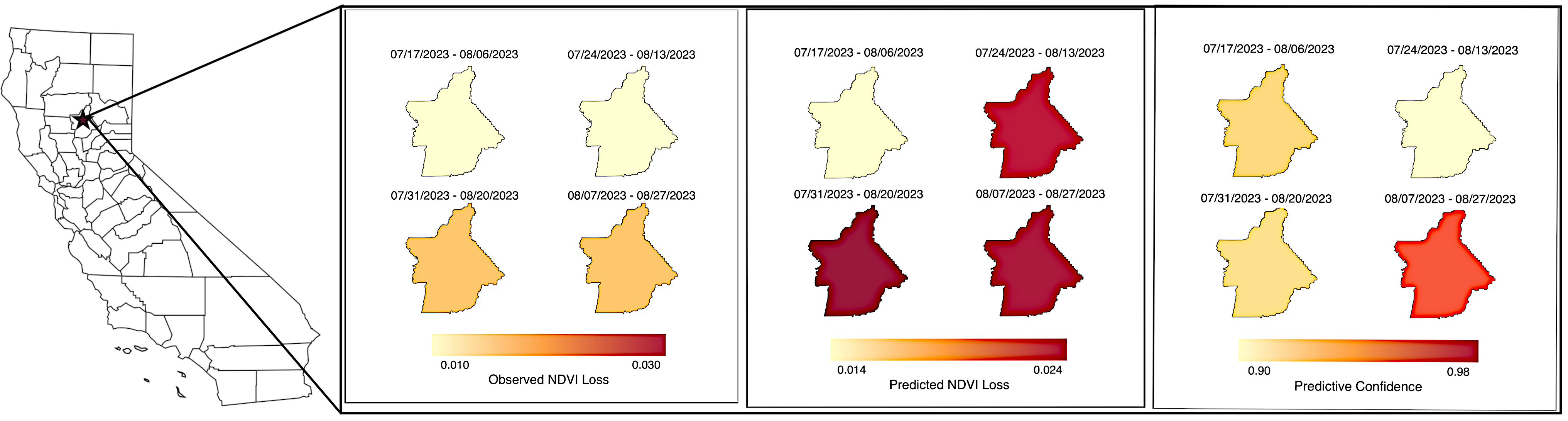}
    \caption{Spatial distribution of MVeLMA's observed, predicted, and predictive-confidence estimates across Butte, California, for four overlapping three-week intervals from July 17 to August 27, 2023. The inset on shows the county’s location within California. The darker colors represent higher vegetation loss and predictive confidence, respectively.}
    \Description{Visual map of California county, Butte county, showing Spatial distribution of MVeLMA's observed, predicted, and predictive-confidence estimates. Darker colors indicate higher values.}
    \label{fig:butte}
\end{figure*}
For temporal analysis, we selected Butte County due to its high frequency of fire events and seasonal vegetation variability. Fire events in Butte County were grouped into overlapping three-week periods, with each period starting one week after the previous period, based on fire start dates. During the initial period (July 17–August 6), Butte County exhibited an average OPFVL of 0.0123, with a predicted PPVL of 0.0091 and a corresponding APC of 0.984, as shown in Figure \ref{fig:butte}. In the subsequent period (August 7–August 27), both OPFVL and PPVL increased to 0.0265 and 0.0248 respectively, while APC remained above 0.97 throughout the season.

The results of this case study demonstrate MVeLMA’s ability to capture both spatial and temporal dynamics of post-fire vegetation loss, while maintaining consistently high predictive confidence across diverse fire conditions.

\section{Conclusion and Future Work}
In this paper, we propose a novel end-to-end probabilistic model called MVeLMA for forecasting county-level vegetation loss from fire events using a multimodal feature integration pipeline and a stacked ensemble-based architecture. To the best of our knowledge, this is the first work to combine the spatiotemporal learning capabilities of BiLSTM, the uncertainty quantification of GPR, and the residual correction strength of RF. MVeLMA model can be used for spatiotemporal prediction of vegetation loss in states with 30-day lead times. Extensive experiments and comparisons to common spatiotemporal models, as well as SOTA models, demonstrate the effectiveness and spatiotemporal predictive capabilities of our proposed model. Furthermore, ablation studies showcase the effectiveness of each individual component of the model. The integration of prediction uncertainty makes it applicable in real-world scenarios as demonstrated by the case study done on wildfire-affected counties of California. 

This work can be further extended in the future by considering the impact of point of interest data, such as landmarks and human-centered activities. Future work should also explore calibration techniques to better align predicted uncertainty with actual prediction errors to handle potential overconfidence. Lastly, incorporating such anthropogenic data into the model can improve the model's accuracy and robustness and provide additional insight for agricultural planning and ecosystem management.

\section{Acknowledgements}
Approved for Public Release; Distribution Unlimited. Public Release Case Number 25-1835. The author’s affiliation with The MITRE Corporation is provided for identification purposes only and is not intended to convey or imply MITRE’s concurrence with, or support for, the positions, opinions, or viewpoints expressed by the author.

\bibliographystyle{ACM-Reference-Format}
\bibliography{references}

\end{document}